\documentclass[preprint]{elsarticle}

\usepackage{hyperref}

\usepackage{amsmath}
\usepackage{tabularx}
\usepackage{multirow}
\usepackage[textwidth = 18cm]{geometry}
\usepackage{multicol}
\journal{Ultrasonics}

\begin{document}

\begin{frontmatter}

\title{Segmentation of Arterial Walls in Intravascular Ultrasound Cross-Sectional Images Using Extremal Region Selection}

\author[uofa]{Mehdi Faraji}
\ead{faraji@ualberta.ca}

\author[uofa]{Irene Cheng}
\ead{locheng@ualberta.ca}

\author[chu]{Iris Naudin}
\ead{iris\_ naudin@yahoo.fr}

\author[uofa]{Anup Basu \corref{cor1}}
\ead{basu@ualberta.ca}

\address[uofa]{Department of Computing Science, University of Alberta, Canada}
\address[chu]{CHU Hospital, Lyon, France}
\address{ \textcopyright 2018. This manuscript version is made available under the CC-BY-NC-ND 4.0 license \url{http://creativecommons.org/licenses/by-nc-nd/4.0/}}
\cortext[cor1]{Corresponding author}

\begin{abstract}
Intravascular Ultrasound (IVUS) is an intra-operative imaging modality that facilitates observing and appraising the vessel wall structure of the human coronary arteries. Segmentation of arterial wall boundaries from the IVUS images is not only crucial for quantitative analysis of the vessel walls and plaque characteristics, but is also necessary for generating 3D reconstructed models of the artery. The aim of this study is twofold. Firstly, we investigate the feasibility of using a recently proposed region detector, namely Extremal Region of Extremum Level (EREL) to delineate the luminal and media-adventitia borders in IVUS frames acquired by 20 MHz probes. Secondly, we propose a region selection strategy to label two ERELs as lumen and media based on the stability of their textural information. We extensively evaluated our selection strategy on the test set of a standard publicly available dataset containing 326 IVUS B-mode images. We showed that in the best case, the average Hausdorff Distances (HD) between the extracted ERELs and the actual lumen and media were $0.22$ mm and $0.45$ mm, respectively. The results of our experiments revealed that our selection strategy was able to segment the lumen with $\le 0.3$ mm HD to the gold standard even though the images contained major artifacts such as bifurcations, shadows, and side branches. Moreover, when there was no artifact, our proposed method was able to delineate media-adventitia boundaries with $0.31$ mm HD to the gold standard. Furthermore, our proposed segmentation method runs in time that is linear in the number of pixels in each frame. Based on the results of this work, by using a 20 MHz IVUS probe with controlled pullback, not only can we now analyze the internal structure of human arteries more accurately, but also segment each frame during the pullback procedure because of the low run time of our proposed segmentation method.
\end{abstract}


\end{frontmatter}

\begin{multicols}{2}

\section{Introduction}
Catheter-based Intravascular Ultrasound (IVUS) has captured considerable attention in the last two decades. This worldwide attention is mostly due to the ability of the imaging method to picture the inside of the human coronary arteries and, hence, provide an opportunity to diagnose and treat cardiovascular diseases such as atherosclerosis (e.g., thin-cap fibroatheroma) that causes a heart attack and a brain stroke \cite{frostegaard2005sle}. Aside from this, the IVUS technique can be helpful in visualizing some internal structures of the human coronary such as the lumen, and thickness and distribution of the plaques \cite{katouzian2012state}. Therefore, IVUS is regularly used to locate the atherosclerosis lesions in the coronary arteries to study the lumen and plaque dimensions, and to guide intervention and stent deployment \cite{ma2016review}.

A typical IVUS imaging system consists of four parts: catheter, transducer, pullback device, and scanning console. The catheter is composed of a 150 cm long guidewire and a tip of 1.2-1.5 mm in size. It is usually inserted in the femoral artery and proceeds toward the coronary arteries. The catheter is responsible for carrying the ultrasound transducer, or other necessary devices, such as inflatable balloons and stents \cite{katouzian2012state}. The transducer is a miniaturized ultrasound probe that emits ultrasound pulses and listens for the backscattered signal. After the catheter has reached the distal end of the coronary, it needs to be manually or automatically pulled back. The speed of the pullback varies between 0.5-1 mm/s \cite{katouzian2012state}. The scanning console is essentially a computer used to post-process the acquired signals (using amplification, filtering, etc.) to provide a user-friendly environment for the surgeon to control the device.

Segmentation of the acquired IVUS images is among the most challenging tasks in medical image analysis. In particular, delineating the interior (lumen) and exterior (media) vessel walls is problematic due to the presence of various artifacts such as motion of the catheter after a heart contraction, guide wire effects, bifurcation and side-branches or similar echogenicity between the vessel wall and some plaques. In some cases even the difference in transducer frequencies affect the segmentation results \cite{katouzian2012state}.

The intrinsic difficulty of IVUS segmentation has attracted many researchers to study and develop solutions using different methodologies, such as intensity-based, statistics and probability-based, active contour and graph search-based approaches. In addition, several methods have been proposed to segment either the lumen or the media or both. A great number of approaches in the literature have utilized the 2D information provided as cross-sectional frames to segment the lumen and media. These 2D cross-sectional gray-scale images are formed after digitization of the backscattered RF signals and are called IVUS B-mode frames. To the best of our knowledge, recent approaches have mostly worked on the B-mode frames which will be reviewed in the following paragraph. For more in-depth reviews of the methods published before 2013, please refer to \cite{katouzian2012state, balocco2014standardized}.    

The lumen and media-adventitia border variations have been modeled within a shape space in \cite{unal2008shape}. The lumen segmentation is then performed by maximizing a nonparametric probability density energy. Also, the edge information has been used to segment the media-adventitia. A physics-based model of the IVUS signal scattered by the structure of the vessel has been used in \cite{mendizabal2016physics} to estimate the differential backscattering cross-sections from the IVUS RF signal. The segmentation curve is obtained after training a Support Vector Machine (SVM) model using the annotated data. Deformable models have been used in \cite{taki2008automatic,zhu2011snake,mendizabal2013segmentation} to detect the border of lumen/intima and media/adventitia. In \cite{taki2008automatic} anisotropic diffusion followed by an edge detector are used to create an initial segmentation which is then corrected using both geometric and parametric deformable models. In \cite{zhu2011snake} an improved version of gradient vector flow (iGVF) has been proposed which includes a balloon force in the snake model that lets the contour pass over leaks and bifurcations. A probabilistic method that formulates the deformation of a lumen contour curve and can be minimized has been proposed in \cite{mendizabal2013segmentation}. However, every first frame of the sequence needs user interactions to manually segment the lumen and media. Following this, a SVM is trained over the annotated data to compute the probability that each pixel is blood or non-blood. A fourfold algorithm based on a deterministic statistical strategy for segmenting the media has been proposed in \cite{zakeri2017automatic}. Their method consists of preprocessing, initial contour detection, active contour segmentation, and contour refinement. First, a sparse binary image is constructed using the local appearance model and the initial contour is elicited. To achieve this, a feature vector is built for each pixel. It includes  gray-level values of the pixel's neighbours, the average intensity of neighbours and the gray-level values of the pixel's neighbours in a contour-enhanced version of the image\cite{zakeri2017automatic}. The K-SVD method is utilized to classify the extracted feature vector. The initial contour is then refined. Next, an active contour model is used to delineate the media border in polar coordinates. The detected contours are then refined using the information provided by identifying the calcification and shadow regions. Artificial Neural Networks have been employed in \cite{su2016artificial} to represent the spatial and neighbouring features of the IVUS image data. As a result, two different vascular structures for lumen and media are extracted and optimized using two ANNs. The borders obtained are then refined and smoothed by an active contour model. In \cite{yan2017novel} the lumen is segmented by a combination of image gradient and fuzzy connectedness model and the media-adventitia border is extracted by a fast marching model. A sequential forward selection process using SVMs and PR curve has been employed to conduct an in-depth analysis of several image features in \cite{vercio2016assessment}. It has been shown that the median filtered image and Haralick's texture features \cite{haralick1979statistical} provide stronger discrimination capabilities for arterial structures. A limitation of their analysis is that it only works for artifact-free IVUS sequences.

As we can see, most approaches have employed either a type of energy minimization method or require annotated data in order to train a classifier or an ANN. However, in this paper we propose a straightforward approach that not only does not require training but also does not use any variational method or deformable model. We show that by extracting EREL features \cite{faraji2015erel,faraji2015extremal} the problem of the IVUS segmentation can be relaxed to a region selection. In particular, we illustrate that it is very likely to find regions similar to lumen and media among the extracted ERELs. Therefore, we propose a selection procedure that efficiently chooses ERELs that are most similar to lumen and media. 

The rest of this paper is organized as follows. In Section \ref{sec::method} we present our proposed method and describe the sequence of Intravascular Ultrasound images that we use throughout this paper. Section \ref{sec::results} illustrates the segmentation results of our proposed method. In Section \ref{sec::discussion} we discuss the advantages and weaknesses of the proposed method. Finally, the concluding remarks are given in Section \ref{sec::conclusion}. 

\section{Materials and Method} \label{sec::method}
\subsection{Materials}
Our proposed method has been evaluated on the test set of publicly available dataset consisting of 326 in-vivo pullbacks of the human coronary artery frames that were acquired by the Si5 (Volcano Corporation), equipped with a 20 MHz Eagle Eye monorail catheter \cite{balocco2014standardized}. The dataset includes a multi-frame 3D context that has between 20 to 50 gated frames acquired using a full pullback at the end-diastolic cardiac phase from 10 patients. Manual annotations for IVUS images are available in the dataset. The annotations have been provided by four clinical experts who work regularly with IVUS echograph. The experts were not aware of other expert's manual annotations and two of them repeated the annotations after about one week from their first delineation \cite{balocco2014standardized}.

The test set contains several types of common artifacts. Specifically, it includes 44 images containing bifurcation, 93 images with a side vessel artifact, and 96 images that have been contaminated by a shadow artifact (some images contain more than one artifact). There are also 143 images that do not contain any serious artifacts except for plaque.
\begin{figure*}[t]
	\setlength\tabcolsep{2pt}
	\centering
	\begin{tabular}{cccc}	
		(a) & (b) & (c) & (d)\\				
		\includegraphics[width=0.2\textwidth]{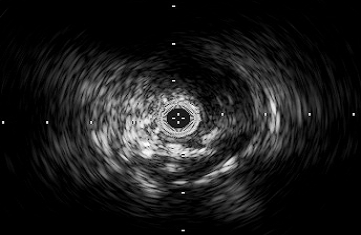} &
		\includegraphics[width=0.2\textwidth]{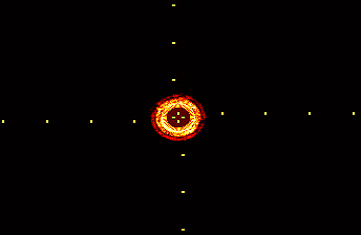} &
		\includegraphics[width=0.2\textwidth]{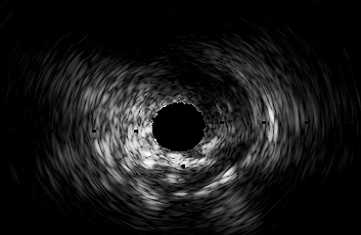} &
		\includegraphics[width=0.33\textwidth]{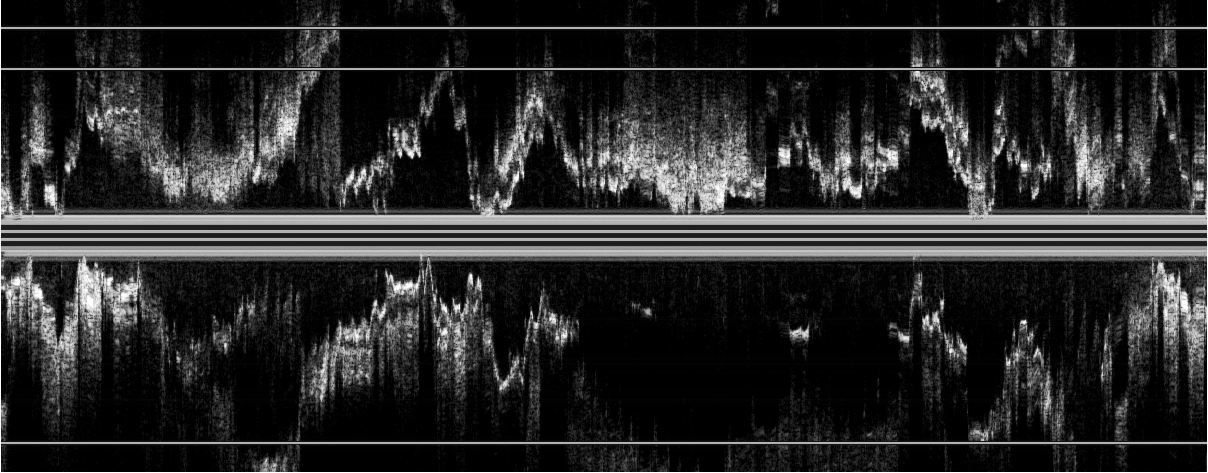} \\  
		(e) & (f) & (g) & (h)\\	
		\includegraphics[width=0.2\textwidth]{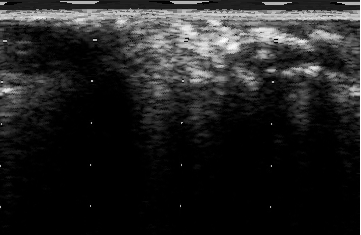} &
		\includegraphics[width=0.2\textwidth]{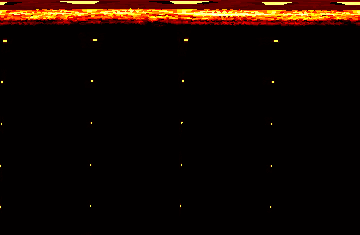} &
		\includegraphics[width=0.2\textwidth]{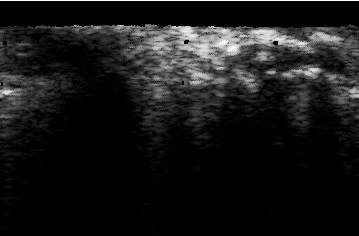} &	
		\includegraphics[width=0.33\textwidth]{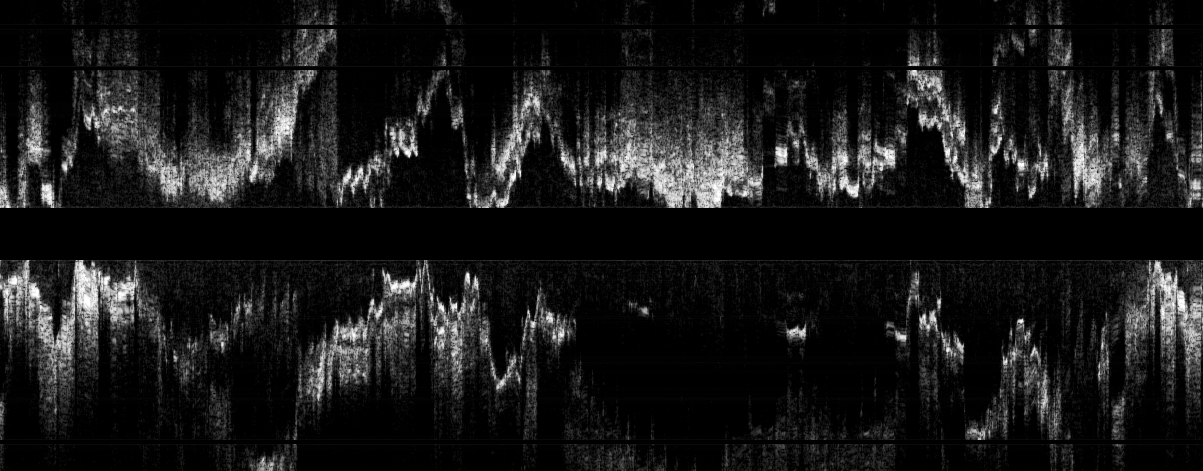} \\			  
	\end{tabular}
	\caption{Artifact removal in B-mode and polar frames. a) A 40 MHz IVUS B-mode frame. b) Computed minimum image of (a). The yellow colour demonstrates higher values and the red colour represents lower values. c) Result of the IVUS frame shown in (a) after the artifact removal. d) A longitudinal cut of the whole volume. The horizontal lines are the effects of the artifact revealed after cutting. e) Corresponding polar frame of (a). f) Calculated minimum image of (e). The yellow colour demonstrates higher values and the red colour represents lower values. g) Corresponding polar frame of (e) after artifact removal. h) Result of artifact removal in all frames of the volume illustrated in a longitudinal view that is cut by the same plane as the one used to cut (d).}
	\label{fig::removeArtifacts} 
\end{figure*}

\subsection{Proposed Method}
In this paper we present a segmentation approach for 20 MHz Intravascular Ultrasound images based on a region detection strategy. Particularly, we investigate whether a recently proposed novel feature extraction method called Extremal Regions of Extremum Levels (EREL) \cite{faraji2015erel,faraji2015extremal} can segment the most essential regions of interest (lumen and media) from the IVUS images required to establish the atherosclerotic plaque area \cite{destrempes2014segmentation}. The proposed method consists of four steps. We first remove the typical artifacts of IVUS frames, such as ring-down effects and calibration squares. Then ERELs are extracted and the obtained regions are filtered based on their types. Next, we perform a region selection procedure to specify two regions as lumen and media. Finally, the contour of the two selected regions is traced and smoothed by an ellipse fitting algorithm.

\subsection{Preprocessing}
Most of the IVUS images have been contaminated by speckle noise \cite{balocco2014standardized}. Speckle is a multiplicative noise that imposes difficulties in processing the Ultrasound images \cite{loizou2008despeckle}. Therefore, to decrease the sensibility of our method to speckle noise, we first use a non-linear median operator to filter the IVUS images.  

One of the main identifiable artifacts in IVUS images is the ring-down effects of the catheter that need to be eliminated from the B-mode frame or from its polar image. Otherwise, there is a high risk of obtaining an erroneous segmentation. To remove the ring-down effects of the catheter we employ the method proposed in \cite{unal2008shape} which is a very fast and straightforward procedure. Detecting the ring-down artifact can be done by processing the whole volume since the artifact is almost available in all of the IVUS frames. Therefore, taking the minimum over all the frames generates an image where there is a significant contrast between the artifact and the non-artifact pixels.

\begin{equation}\label{eq::minImage}
I_{min}(x,y) = \min_{i \in \lambda} I_i(x,y)
\end{equation}

where $\lambda$ is a set of available frames in a particular IVUS sequence. We can then locate the artifacts' coordinates by subtracting the artifact zones from every frame, as in \cite{unal2008shape}. Figure~\ref{fig::removeArtifacts}(b) and Figure~\ref{fig::removeArtifacts}(f) show the resulted minimum image in B-mode and polar frame respectively.

Another type of artifact that we can detect using Eq.\ref{eq::minImage} is the calibration square artifact. These small squares have a very bright constant intensity in all frames that remains bright in the minimum image. In several longitudinal cuts of the IVUS volume the effects of these artifacts are revealed as horizontal lines. Figure~\ref{fig::removeArtifacts} illustrates both the artifacts and the resulting image after removing them. 

\begin{figure*}[t]
	\setlength\tabcolsep{2pt}
	\centering
	\begin{tabular}{cccc}					
		\includegraphics[width=0.23\textwidth]{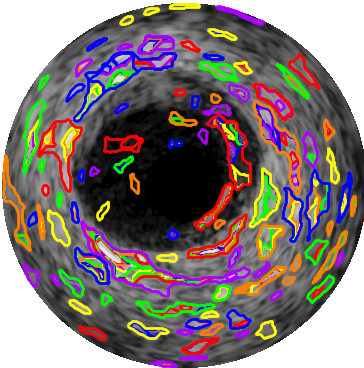} &
		\includegraphics[width=0.23\textwidth]{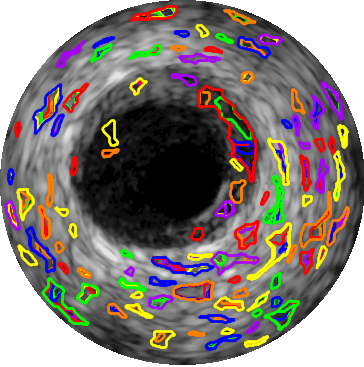} &			
		\includegraphics[width=0.23\textwidth]{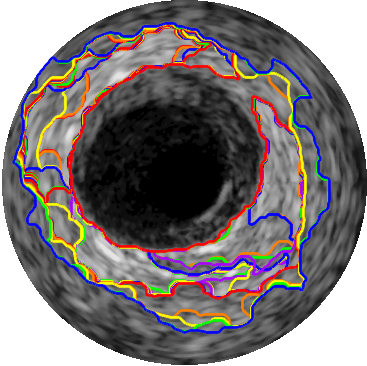} &
		\includegraphics[width=0.23\textwidth]{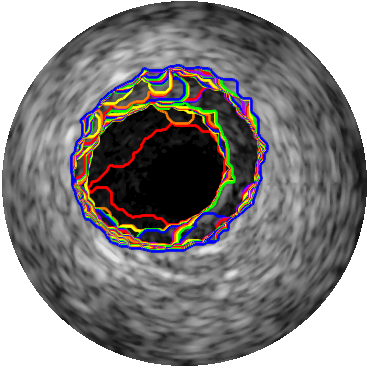} \\
		(a) & (b) & (c) & (d) \\
	\end{tabular}
	\caption{Extracted ERELs from a 20 MHz IVUS frame belonging to dataset \cite{balocco2014standardized}. The initial parameters of EREL are: $\alpha = 0.5$, $\beta = 1$, $A_{min} =(R\times C)/100 = 1474$ and $A_{max} =(R\times C)/3 = 49152$.  a) $Q^-$ regions with small area. b) $Q^+$ regions with small area. c) $Q^-$ regions with large area. d) $Q^+$ regions with large area. Contour colours have been randomly assigned and are only for visualization purposes.}
	\label{fig:erels} 
\end{figure*}

\subsection{Extremal Regions of Extremum Levels}
EREL\cite{faraji2015erel,faraji2015extremal} is a region detector that employs a union-find structure \cite{Sedgewick11} in conjunction with the edge information to detect a series of connected pixels from the image. The edge information of the image is included in the method by using the Maxima of Gradient Magnitude (MGM) points. The idea underlying EREL is to binarize the image with all possible integer thresholds and analyze the results obtained based on their global criterion and their local edge information. The regions belonging to the globally distinguished levels (Extremum Levels) are then extracted from the union-find tree. 

Generally, two types of regions can be extracted from a gray-level image. The first type includes regions that evolve from brighter surfaces to darker boundaries which are identified by $Q^{-}$. The superscript '$^{-}$' emphasizes the fact that the intensity values are decreased from the surface of the regions towards the boundaries. The second type consists of regions that evolve from darker surfaces to brighter boundaries and are denoted by $Q^+$. This type of the region is consistent with the inherent characteristics of the lumen and media visualized by the backscattered 20MHz IVUS signals. Therefore, we only need to extract $Q^+$ regions to obtain the lumen and media because both regions evolve from darker surfaces to brighter boundaries.

To use EREL, we need to set several initialization parameters, namely $A_{min}$, $A_{max}$, $\alpha$, $\beta$. These parameters define the functionality of the detector and can be tuned based on the application \cite{faraji2015extremal,faraji2015erel}. In particular, we use $A_{min}$, $A_{max}$ to set the minimum and maximum area of the extracted regions. To better separate small regions from bigger ones, we choose a value for $A_{min}$ that correlates with image dimensions. Specifically, we set $A_{min} = (R\times C)/100$ and $A_{max} = (R\times C)/3$ where $R$ and $C$ represent the number of rows and columns of the IVUS image, respectively. The parameter $\alpha$ is usually in $[0~2.5]$ and represents the strength of the resulting interest points \cite{faraji2015extremal}. In this study we set $\alpha = 0.5$. Finally, $\beta =1$ denotes the width of the moving window over the global criterion vector \cite{faraji2015extremal}. The extracted EREL regions using the above-mentioned values are illustrated in Figure~\ref{fig:erels}.

However, not all four types of the regions depicted in Figure~\ref{fig:erels} encompass lumen and media regions. In fact, we only need to extract large area $Q^+$ regions as illustrated in Figure~\ref{fig:erels}(d). Since large area $Q^+$ regions contain the actual lumen and media segments, detectors need not track $Q^-$ regions and, therefore, omit unnecessary computations which eventually helps to have a faster detector.

\subsection{EREL Selection}\label{sec::erelSelection}
The goal of this section is to address the problem of finding the most appropriate ERELs to be designated as lumen and media. By the most appropriate we mean the closest regions to the gold standard. As can be seen in Figure~\ref{fig:erels}(d), although ERELs are nested regions, it is clear that there is at least one EREL that is very close to the true lumen and similarly there is at least one EREL that corresponds to the true media. Therefore, we can relax the problem of lumen and media segmentation to only a selection procedure, i.e., assigning two nested ERELs to lumen and media. 

Local maxima searching is the approach that we employ to select lumen and media from the nested set of ERELs. Assume that we have a vector (denoted by $V$) representing the evolution of the $Q^+$ regions. The index of each element of the vector $V$ corresponds to a $Q^+$ region. Since the extracted ERELs start from the smallest enclosing region and end with the largest enclosing region, the actual region corresponding to the lumen should be found among the regions located at the early indices of the vector. Likewise, a region representing the media is expected to be found among the regions belonging to the end section of the vector $V$. 

Our aim is to construct a vector that ultimately gives us the stability of the regions in terms of the length of the boundary, average intensity and entropy variation. As we can see in Figure~\ref{fig:erels}(d), the boundaries of the $Q^+$ regions are not smooth and are subject to large variations. Therefore, lumen and media regions should be selected among those $Q^+$ regions that have more stable boundary length variations. Also, the average intensity of the $Q^+$ regions should be stable enough (i.e., they should not change much over several subsequent regions). Entropy measure can be used in order to create a feature vector that is sensitive to textural information of the regions. 
\begin{figure*}[t]
	\centering
	\begin{tabular}{cc}
		\includegraphics[width=0.485\textwidth]{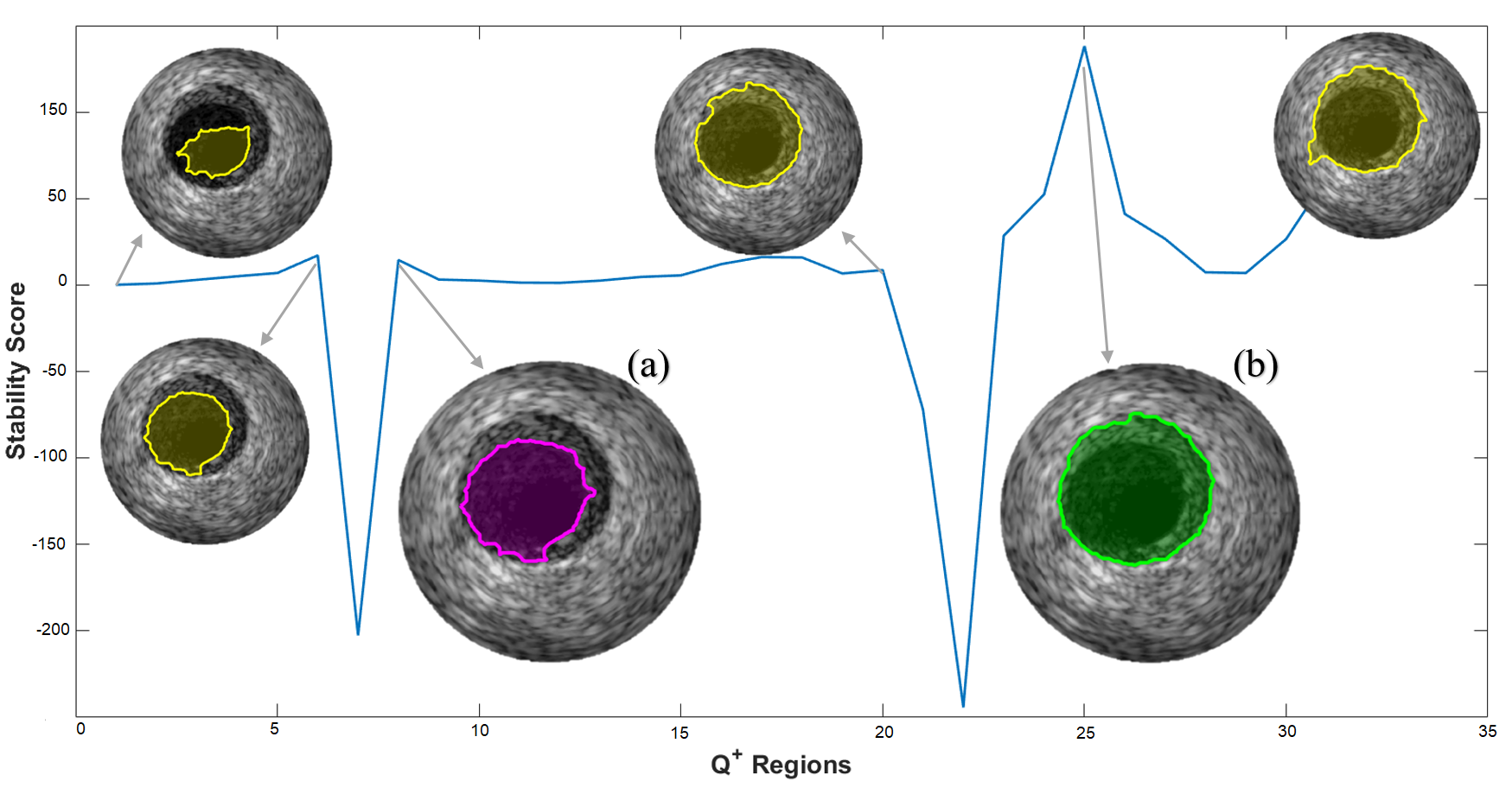} & \includegraphics[width=0.45\textwidth]{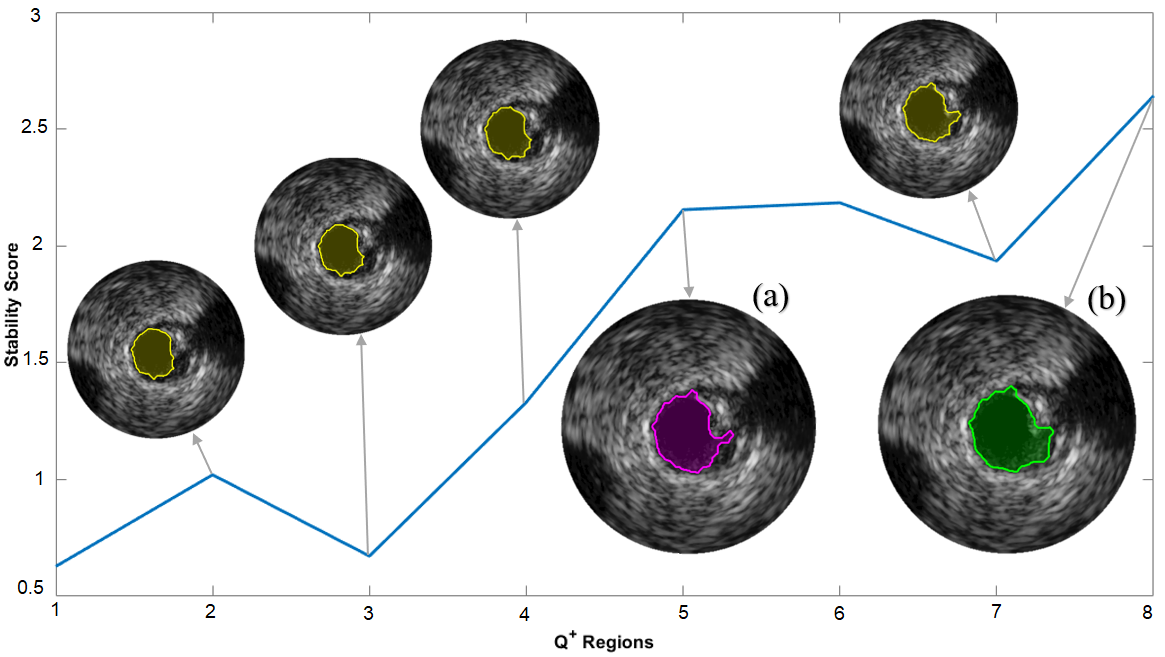}\\
		(i) No Artifact & (ii) Plaque and Shadow\\		
	\end{tabular}	 
	\caption{The evolution of the $Q^+$ regions and their stability criteria in presence of no artifacts vs. plaque and shadow artifact. a) The best candidate region representing the lumen. b) The best region representing the media. The neglected regions are highlighted by the yellow colour and the selected regions for lumen and media are indicated by magenta and green colour.}
	\label{fig::selectErels} 
\end{figure*}
The calculation of these three features are straightforward. The boundary lengths of the regions are available as an output of the EREL algorithm and are calculated based on a bottom-up tracking of boundary pixels along a parametric curve $C(p)$.
\begin{equation}\label{eq::boundaryLength}
\mathcal{L} = \int_{0}^{1} |\dfrac{\partial C(p)}{\partial p}| dp
\end{equation}

Additionally, the calculation of the average intensity of ERELs can be readily done.
\begin{equation}\label{eq::average}
E = \dfrac{\sum_{i = 1}^{N}Q^+_i}{N}
\end{equation}
where $N$ represents the area of the region and $Q^+_i$ is the intensity value of pixel $i$. The entropy measure of a grayscale region \cite{haralick1979statistical} is denoted as follows.
\begin{equation}\label{eq::entropy}
\mathcal{H} = - \sum_i^K p_i \log_2 p_i
\end{equation}
where $p_i$ is the value of the bin $i$ of the normalized histogram of the region and captures the probability of having a pixel with a certain gray-value. $K$ is the number of available bins in the normalized histogram.

Afterwards, for every IVUS image, we create a vector ($V$) where each element is obtained from the product of the above-mentioned measures for each region.
\begin{equation}
V =
\begin{bmatrix}
\mathcal{L}_1 E_1 \mathcal{H}_1 & \mathcal{L}_2 E_2 \mathcal{H}_2 & ... & \mathcal{L}_n E_n \mathcal{H}_n 
\end{bmatrix}
\end{equation}
where $n$ is the number of $Q^+$ regions extracted by the EREL algorithm. 

Vector $V$ illustrates the variation of textural information of regions through the extraction and evolution of $Q^+$ regions. It is strictly increasing because $Q^+$ regions are nested and non-repetitive. More stable sequences of vector $V$ are more likely to represent lumen and media since these stable sequences shows that the extremal regions are subject to saturation and the subsequent regions might contain a noticeable change. The best way to find the stable regions is to create a vector describing the stability score of regions. Every element of the stability vector is calculated as follows.
\begin{equation}\label{eq::omega}
\Omega_i = \dfrac{V_i}{V_{i+1}-V_{i-1}}
\end{equation}
where $i$ specifies a specific element of the vector $V$ and varies from one to the number of the detected $Q^+$ regions for an IVUS image. 

The local maxima of the stability score points to regions with high stability because the ratio of their current value to the change among their two neighbours is larger than the other surrounding elements. So, we select lumen and media from the detected local maxima. Specifically, a  $Q^+$ region with higher prominence value among the first two peaks is considered as lumen. If the IVUS image contains no artifacts, the media will be represented by the last detected peak. Based on our observation, the stability score of the images that contain serious artifacts have none or a small number of peaks since the presence of the artifacts interferes with the natural extraction of $Q^+$ regions (see Figure~\ref{fig::selectErels}(ii)). Therefore, when a small number of local maxima is detected, we consider the last extracted region as media. This process for an IVUS with no particular artifact is illustrated in Figure~\ref{fig::selectErels}. The local maxima of the stability score indicates the regions for which the variation of the textural characteristics is more stable than their surrounding regions.  As can be seen in Figure~\ref{fig::selectErels}(i)(a), the second peak is selected as a suitable region for the lumen since it has a higher prominence than the first peak. Also, the region corresponding to the last peak of $\Omega$ which is chosen as the media has been shown in  Figure~\ref{fig::selectErels}(i)(b).

It is important to note that before searching for the local maxima we need to remove outliers. To find outliers we employ the modified Z-score normalization suggested in \cite{iglewicz1993detect}.
\begin{equation}\label{eq::removeOutliers}
M_i = \dfrac{0.6745(A_i-\tilde{A})}{MAD}
\end{equation}
where $A_i$ represents the area of the region $i$, $\tilde{A}$ is the median of a vector of all region areas, and $MAD$ denotes the Median Absolute Deviation and is calculated by the following equation.
\begin{equation}
MAD = median(|A_i-\tilde{A}|)
\end{equation}
Therefore, the regions that have a modified Z-score less than $Z_{min} = -3$ and greater than $Z_{max} = 3$ are unlikely to represent lumen and media and hence can be removed from the selection process.
\subsection{Contour Extraction}
Since the general shape of the lumen and media regions of the vessel are very similar to conic sections, we propose to represent lumen and media by ellipses. To find the pixels inside and on the ellipse border, it is sufficient to find the orientation of the ellipse, the major and minor axis length. Generally, EREL outputs all pixels belonging to each extracted region in addition to its shape description parameters which are three coefficients of ellipse equation \cite{faraji2015erel,faraji2015extremal}. Specifically, if a region $Q$ is described by an ellipse with three coefficients, namely $c$,$d$ and $e$ then: 
\begin{equation}
\forall x,y \in Q : cx^2 + 2dxy + ey^2 \le 1
\end{equation}
There is a direct relationship between the parameters of the fitted ellipse and the second central moments \cite{shapiro2001computer}:
\begin{equation}
M = \begin{bmatrix}
c & d \\
d & e
\end{bmatrix} =
\dfrac{1}{4(\mu_{yy}\mu_{xy}-\mu_{xx}^2)}
\begin{bmatrix}
\mu_{yy} & \mu_{xy}\\
\mu_{xy} & \mu_{xx}
\end{bmatrix}
\end{equation}
where $\mu_{xx}$, $\mu_{xy}$ and $\mu_{yy}$ are the second order central moments and are calculated as follows \cite{shapiro2001computer}:
\begin{equation}
\mu_{xx} = \dfrac{1}{A}\displaystyle\sum_{(x,y)\in Q} (x-\bar{x})^2
\end{equation}
\begin{equation}
\mu_{yy} = \dfrac{1}{A}\displaystyle\sum_{(x,y)\in Q} (y-\bar{y})^2
\end{equation}
\begin{equation}
\mu_{xy} = \dfrac{1}{A}\displaystyle\sum_{(x,y)\in Q} (x-\bar{x})(y-\bar{y})
\end{equation}
where $A$ represents the area of the region and $(\bar{x},\bar{y})$ specifies the coordinates of the region's centroid.

After having obtained matrix $M$ (reported by EREL), finding pixels belonging to ellipse border of the regions is straightforward. We assume that $M = \mathbf{V} \mathbf{\Lambda} \mathbf{V}^{-1}$ where $\mathbf{V}$ denotes a matrix containing the eigenvectors and $\mathbf{\Lambda}$ indicates a matrix of eigenvalues.  The minimum ($\lambda_{min}$) and maximum ($\lambda_{max}$) eigenvalues in $\mathbf{\Lambda}$ are used to calculate the length of the minor($b$) and major ($a$) axes of the ellipse, respectively \cite{szeliski2010computer}.
\begin{equation}
a = \dfrac{1}{\sqrt{\lambda_{max}}}
\end{equation}
\begin{equation}
b = \dfrac{1}{\sqrt{\lambda_{min}}}
\end{equation}
The orientation of the ellipse ($\theta$) can be recovered by calculating the angle between the major axis and the $x$ axis. Let $\mathbf{V} = [\mathbf{u}~\mathbf{v}]$ where $\mathbf{u} = [u_1~u_2]^T$ and $\mathbf{v} =[v_1~ v_2]^T$ are two column vectors corresponding to the minimum and maximum eigenvalues, respectively. The orientation of the ellipse is then obtained as follows.
\begin{equation}
\theta = \tan^{-1}\dfrac{v_1}{v_2}
\end{equation}

\subsection{Computational Cost}
The run time of the proposed method depends only on the complexity of EREL which is $O(N)$ \cite{faraji2015extremal}, where $N$ is the total number of pixels in the image. The subsequent operation proposed in this paper in Section \ref{sec::erelSelection} works on a constant number of regions. The maximum number of nested regions that can be extracted from the same root pixel is at most 256 for an 8-bit image \cite{faraji2015extremal}. In our experiments with IVUS images, the number of candidate ERELs ($Q^+$) rooted from the center of image are even less (lower than 75). Therefore, the region properties denoted in Eq.\ref{eq::boundaryLength}, Eq.\ref{eq::average}, Eq.\ref{eq::entropy} and Eq.\ref{eq::omega} and the second moment matrix are calculated on a constant number of regions containing far less than $N$ pixels. Therefore, the overall run time of the proposed method is $O(N)$ in the worst case. A comparison between the actual run time of the proposed method and the methods reported in \cite{balocco2014standardized} is shown in Table~\ref{tbl::runningTime}. 

\begin{center}
\begin{table*}[!ht]
	\centering \footnotesize
	\caption{Comparison of each method's run time (required for segmenting a frame) reported in \cite{balocco2014standardized} and the proposed method.}
	\begin{tabular}{|l|c|c|c|c|l|}
		\hline 
		                                    & \textbf{Category}        &\textbf{Semi/Auto}& \textbf{2D/3D}& \textbf{Time per frame} &\textbf{ Hardware used}\\
		 \hline
		 Proposed Method                   & Lumen and media & AUTO                          & 2D  & 0.19 s & Core i7-4700HQ, 2.4 GHz\\ 
        Unal et al. \cite{unal2008shape}   & Lumen and media & SEMI & 2D  & 3.25 s & Pentium 6200 2.13 GHz\\
        Wang et al.\cite{balocco2014standardized} & Lumen & SEMI & 2D  & 1 m 40 s  & Xeon 2.67 GHz\\
        Destrempes et al.\cite{balocco2014standardized}  & Lumen and media & SEMI & 2D &8.64 a & Core i7 Q740 @ 1.73 GHz \\
        Downe et al. \cite{downe2008segmentation}  & Lumen and media & AUTO & 3D & 0.16 s & Core 2, 2.4 GHz\\
        Alberti et al.\cite{balocco2014standardized} & Lumen & AUTO & 3D  & 13 s          & Core 2, Duo 2.13 GHz\\
        Ciompi et al. \cite{ciompi2012holimab} & Media          & AUTO   & 2D   & 20 s & Core i7, 2.8 GHz\\
        Mendizabal et al. \cite{mendizabal2013segmentation}   & Lumen & SEMI & 2D  & 4.96 s & Core i7, 2 GHz\\
        Exarchos et al. \cite{balocco2014standardized}        & Lumen and media & AUTO & 2D  & 0.5 s & Core 2, Duo 3.33 GHz\\		
		\hline
	\end{tabular} 
   \label{tbl::runningTime}
\end{table*}	
\end{center}

\section{Results}\label{sec::results}
In this section, we present the segmentation results of our method. Also, by showing the best case results, we demonstrate that regions very close to the lumen and media exist among the extracted ERELs and a proper selection strategy (the proposed method) can distinctively select the lumen and media regions from the extracted ERELs. Furthermore, we present extensive evaluation results of our method based on three standard evaluation metrics on IVUS frames containing various artifacts.

\subsection{Evaluation Measures}\label{sec::eval}
To assess the segmentation obtained by our method, we employ three evaluation metrics, namely Jaccard Measure (JM), Hausdorff Distance (HD), and Percentage of Area Difference (PAD). Using these metrics that have been also used in \cite{balocco2014standardized} to evaluate the results of 8 state-of-the-art methods on the same dataset makes it possible to draw a fair comparison between our method and the results reported in \cite{balocco2014standardized}. 

The Jaccard Measure is calculated based on the comparison of the automatic segmentation result and the manual segmentation delineated by experts. It quantifies the overlap area between the automatic and manual segmentation as computed by the following equation.
\begin{equation}
JM = \dfrac{R_{auto}\cap R_{man}}{R_{auto}\cup R_{man}}
\end{equation}
Where $R_{auto}$ is the vessel region segmented by the method and $R_{man}$ represents the region that has been segmented manually by experts.

The Hausdorff Distance between the automatic ($C_{auto}$) and manual ($C_{man}$) curves is the greatest distance of all points belonging to $C_{auto}$ to the closest point in $C_{man}$ and is defined as follows \cite{molinari2011caudles}.
\begin{equation}
HD = \max \{{d(C_{man},C_{auto}),d(C_{auto},C_{man})}\}
\end{equation}
To calculate $d(C_{auto},C_{man})$ first the minimum of all Euclidean distances from each point belonging to $C_{auto}$ to all points in $C_{man}$ is obtained. Then, $d(C_{auto},C_{man})$ is computed by taking the maximum of all the minimum distances. Similarly, $d(C_{man},C_{auto})$ is computed by taking the maximum of all minimum distances from $C_{man}$ to $C_{auto}$ \cite{molinari2011caudles}. 

The Percentage of Area Difference calculates the segmentation area difference between the automatic ($A_{auto}$) and manual ($A_{man}$) segmentation and is computed as follows.
\begin{equation}
PAD = \dfrac{|A_{auto}-A_{man}|}{A_{man}}
\end{equation}

\begin{table*}\centering \footnotesize
		\caption{The best case performance results of the proposed method. Measures represent the mean and standard deviation (std) evaluated on 435 frames of dataset \cite{balocco2014standardized}. The measures are categorized based on the presence of a specific artifact in each frame. The evaluation measures are Jaccard Measure (JM), Hausdorff Distance (HD), and Percentage of Area Difference (PAD).}
	\begin{tabular}{|c|l|c@{}|c@{}|c@{}|c@{}|c@{}|c@{}|}
		\cline{3-8} 
		\multicolumn{1}{c}{}&\multicolumn{1}{c|}{} & \multicolumn{3}{c}{\normalsize\textbf{Lumen}} \vline& \multicolumn{3}{c}{\normalsize\textbf{Media}} \vline\\
		\cline{3-8}
		\multicolumn{1}{c}{}& \multicolumn{1}{c|}{}&\textbf{HD} & \textbf{JM} & \textbf{PAD} &\textbf{HD} & \textbf{JM} & \textbf{Pad} \\ 
		\hline 
		\multirow{3}{*}{\parbox[c]{1.8cm}{\centering \scriptsize \textbf{General}\\ \textbf{Performance}}} 
		& \textit{EREL}      &  0.22 (0.12) & 0.91 (0.04) & 0.03 (0.03) &  0.50 (0.45) &0.83 (0.15)	    &  0.13 (0.15)\\ 
		& \textit{Intra-obs} &  0.28 (0.13) & 0.88 (0.05) & 0.11 (0.08) &  0.24 (0.12) &0.92 (0.03) 	& 0.06 (0.04)\\
		& \textit{Inter-obs} &  0.17 (0.13) & 0.93 (0.05) & 0.04 (0.06) &  0.14 (0.12) &0.95 (0.03) 	& 0.03 (0.03)\\
		
		\hline
		
		\multirow{3}{*}{\scriptsize\textbf{No Artifact}} 
		& \textit{EREL}      & 0.20 (0.09)& 0.92 (0.03) & 0.02 (0.02)  	& 0.23 (0.17)& 0.92 (0.05) & 0.03 (0.05) \\
		& \textit{Intra-obs} & 0.28 (0.13)& 0.88 (0.05) & 0.11 (0.08)  	& 0.24 (0.12)& 0.92 (0.03) & 0.06 (0.04) \\
		& \textit{Inter-obs} & 0.17 (0.13)& 0.93 (0.05) & 0.04 (0.06)  	& 0.14 (0.12)& 0.95 (0.03) & 0.03 (0.03) \\
		\hline
		
		\multirow{3}{*}{\scriptsize\textbf{Bifurcation}}  
		& \textit{EREL}      & 0.36 (0.22)& 0.85 (0.07) & 0.07 (0.07)  & 0.95 (0.49)& 0.67 (0.16) & 0.27 (0.17) \\ 
		& \textit{Intra-obs} & 0.30 (0.12)& 0.88 (0.04) & 0.09 (0.06)  & 0.24 (0.09)& 0.92 (0.02) & 0.06 (0.03) \\
		&\textit{ Inter-obs} & 0.18 (0.21)& 0.92 (0.07) & 0.05 (0.09)  & 0.15 (0.13)& 0.95 (0.04) & 0.03 (0.03) \\
		
		\hline
		
		\multirow{3}{*}{\textbf{\scriptsize Side Vessels}} 
		& \textit{EREL}      & 0.20 (0.09)& 0.90 (0.04) & 0.03 (0.03) 	& 0.62 (0.53) & 0.78 (0.16) & 0.18 (0.16) \\ 
		& \textit{Intra-obs} & 0.30 (0.13)& 0.88 (0.05) & 0.10 (0.08)   & 0.24 (0.11) & 0.92 (0.04) & 0.06 (0.04) \\
		&\textit{ Inter-obs} & 0.20 (0.11)& 0.91 (0.05) & 0.06 (0.05)   & 0.15 (0.10) & 0.95 (0.03)	& 0.03 (0.04) \\
		
		\hline
		
		\multirow{3}{*}{\textbf{\scriptsize Shadow}}  
		& \textit{EREL}      & 0.22 (0.11)& 0.89 (0.04) & 0.03 (0.03) 	& 1.10 (0.38)& 0.64 (0.13) & 0.31 (0.14) \\ 
		&\textit{ Intra-obs} & 0.31 (0.13)& 0.88 (0.05) & 0.11 (0.08)  	& 0.27 (0.15)& 0.92 (0.04) & 0.06 (0.05) \\
		& \textit{Inter-obs} & 0.18 (0.14)& 0.93 (0.05) & 0.04 (0.06)  	& 0.14 (0.10)& 0.96 (0.03) & 0.02 (0.02)\\
		
		\hline
	\end{tabular} 
	\label{tbl::bestcase}
	
\end{table*}

\subsection{Best Case Results}
In order to show that the extracted EREL regions have the potential to represent lumen and media regions, we evaluate all of the extracted ERELs by calculating the evaluation metrics (Section \ref{sec::eval}) for the contours of each EREL and the manually annotated contours. Then, the ERELs that correspond to the maximum JM of lumen and media are selected as the best extracted ERELs for that frame. The quantitative results in comparison with the intra-observer and inter-observer variability are reported in Table~\ref{tbl::bestcase}.

\begin{figure*}
	\setlength\tabcolsep{2pt}
	\centering
	\begin{tabular}{ccccc}	
		\includegraphics[width=0.18\textwidth]{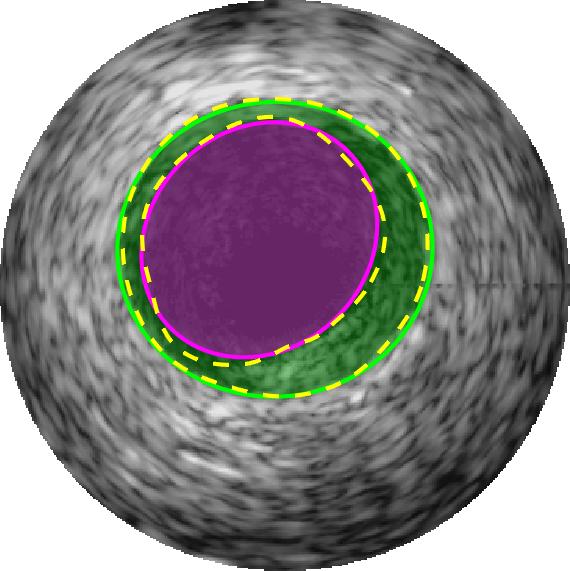} &
		\includegraphics[width=0.18\textwidth]{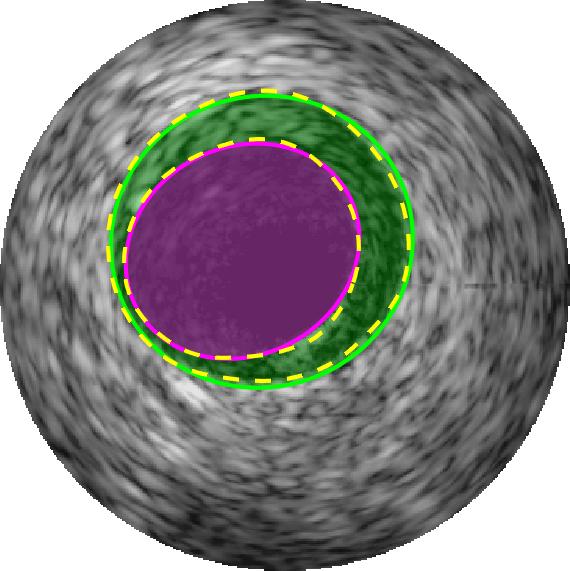} &
		\includegraphics[width=0.18\textwidth]{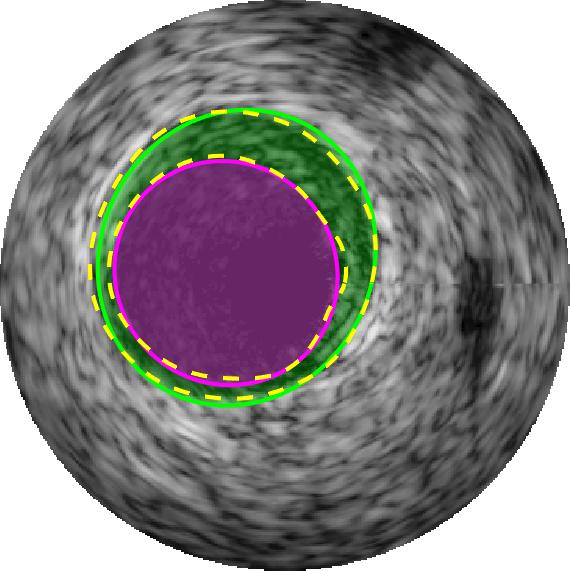} &
		\includegraphics[width=0.18\textwidth]{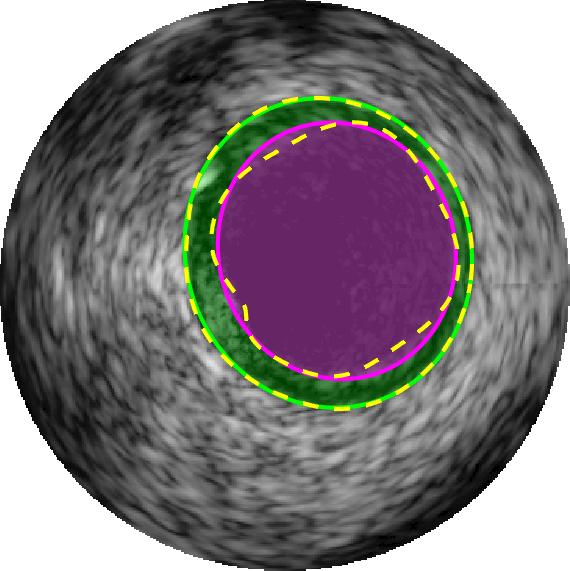} & 
		\includegraphics[width=0.18\textwidth]{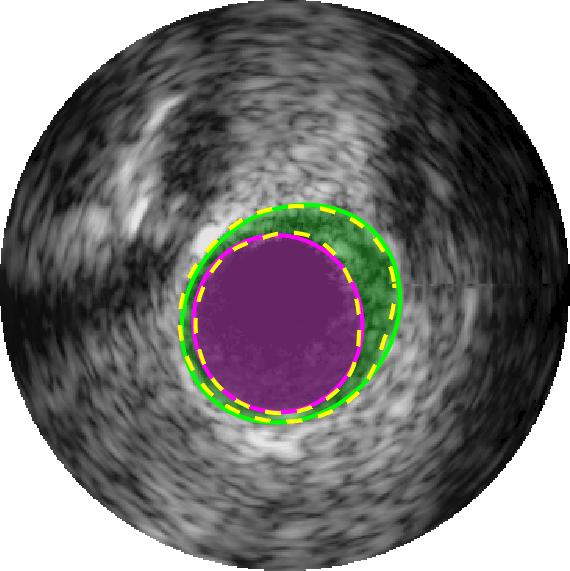} \\
		\includegraphics[width=0.18\textwidth]{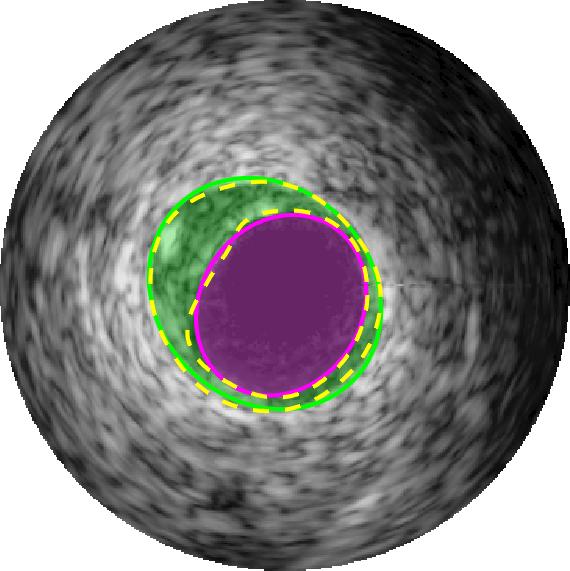} &
		\includegraphics[width=0.18\textwidth]{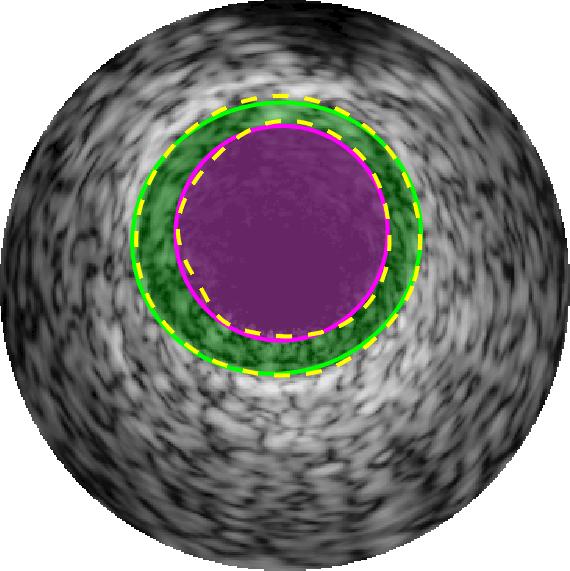} &	
		\includegraphics[width=0.18\textwidth]{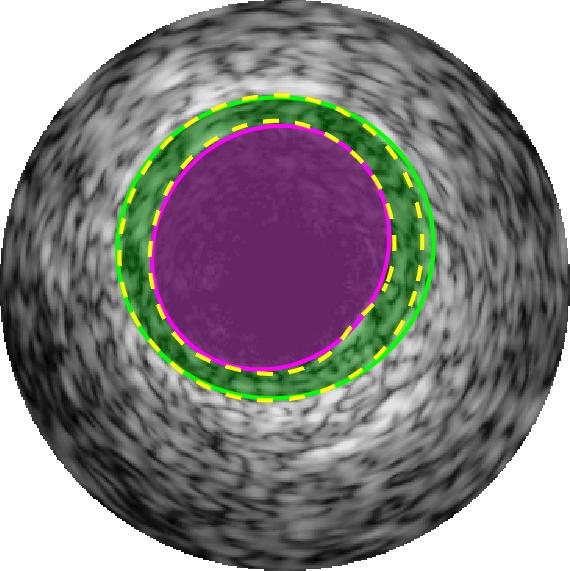} &			  
		\includegraphics[width=0.18\textwidth]{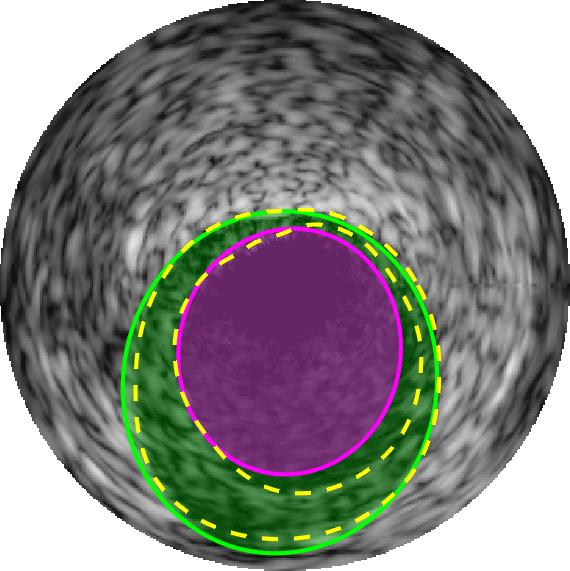} &
		\includegraphics[width=0.18\textwidth]{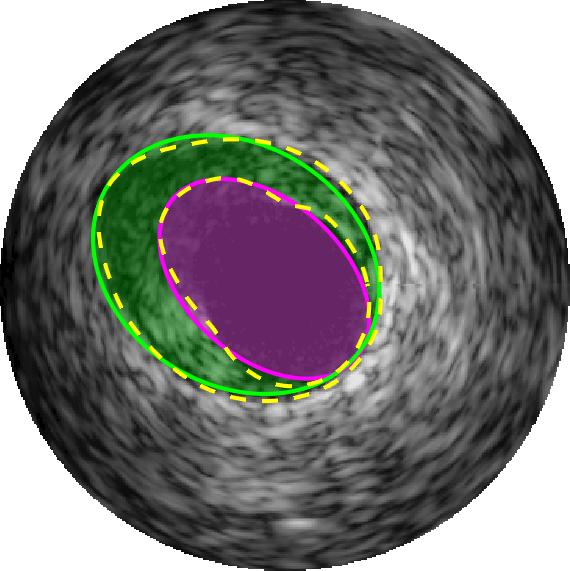} \\
		\includegraphics[width=0.18\textwidth]{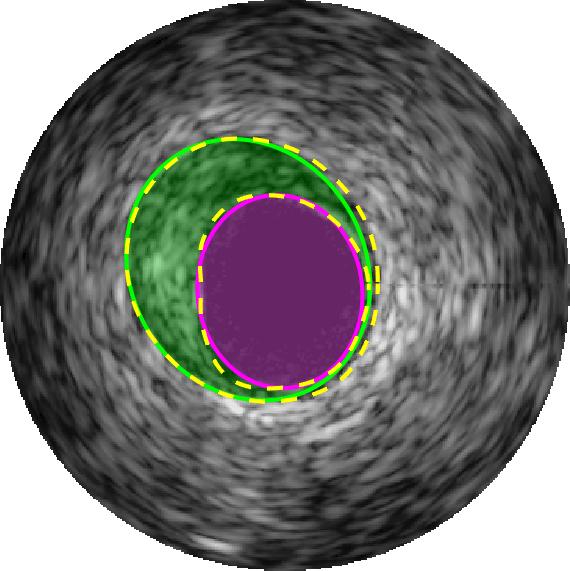} &
		\includegraphics[width=0.18\textwidth]{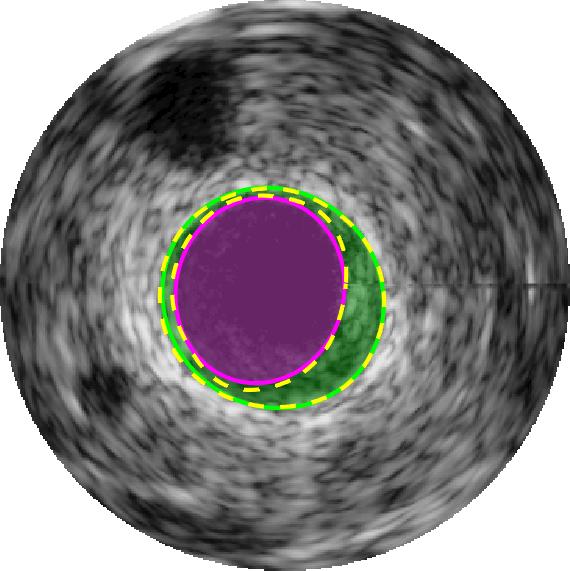} &
		\includegraphics[width=0.18\textwidth]{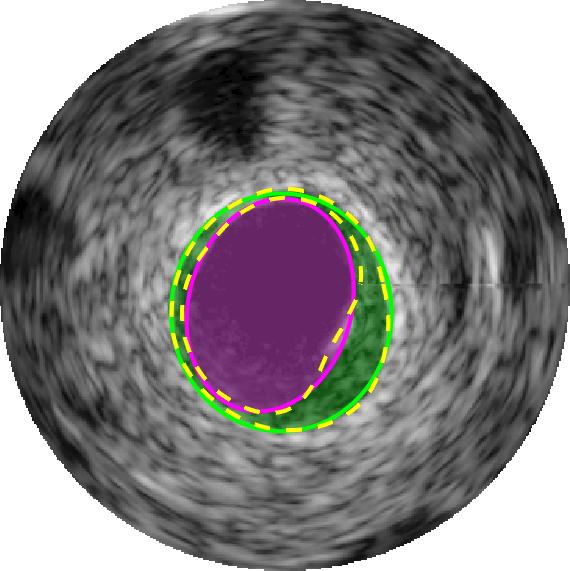} &
		\includegraphics[width=0.18\textwidth]{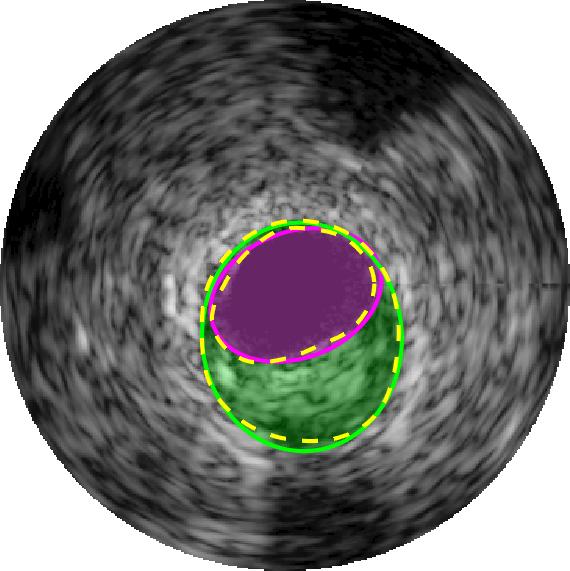} &
		\includegraphics[width=0.18\textwidth]{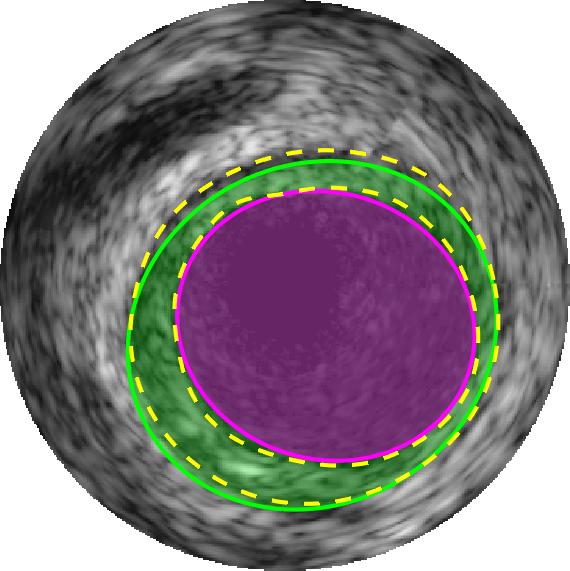} \\	
		\includegraphics[width=0.18\textwidth]{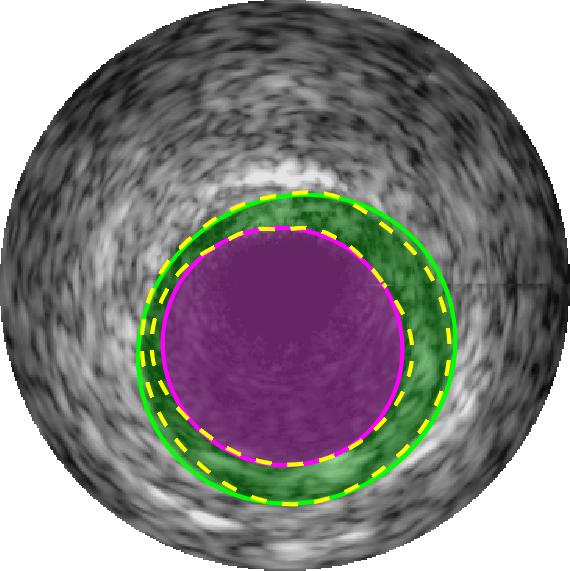} &			  
		\includegraphics[width=0.18\textwidth]{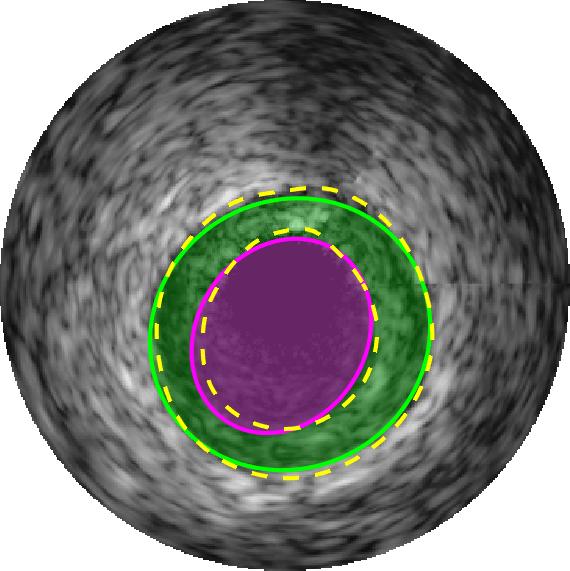} &
		\includegraphics[width=0.18\textwidth]{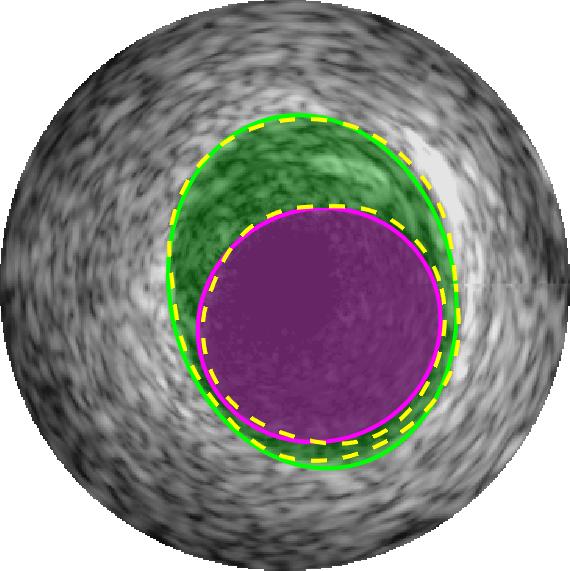} &
		\includegraphics[width=0.18\textwidth]{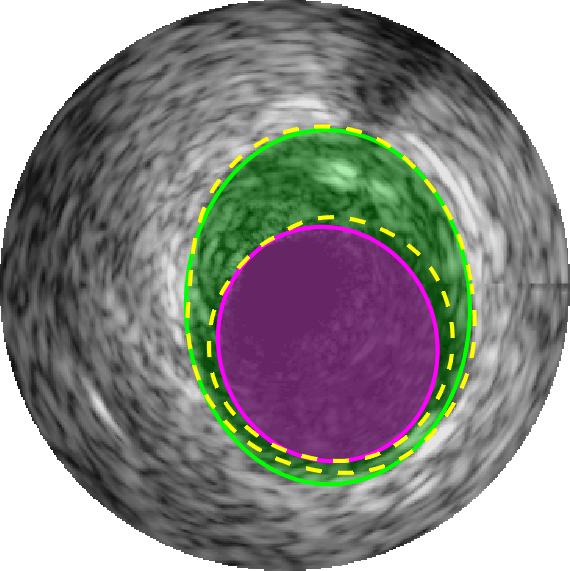} &
		\includegraphics[width=0.18\textwidth]{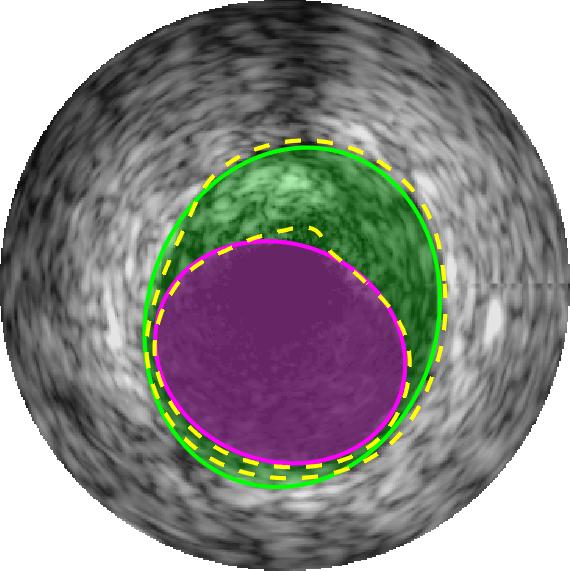} \\
		\includegraphics[width=0.18\textwidth]{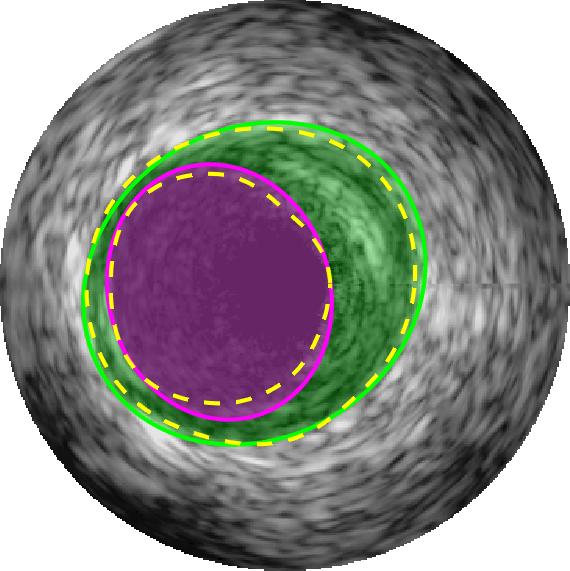} &		
		\includegraphics[width=0.18\textwidth]{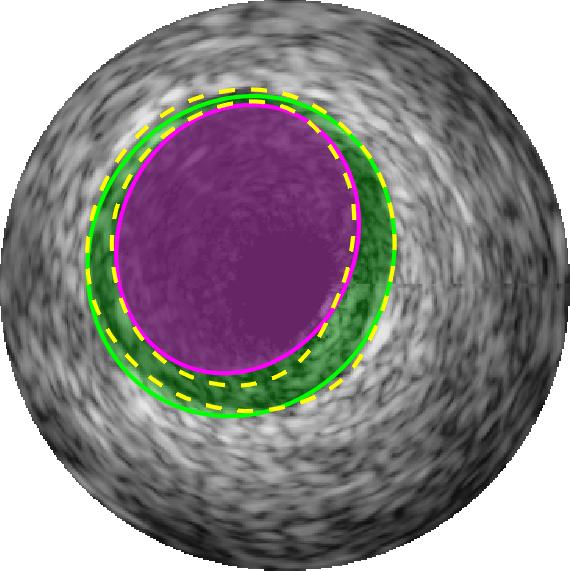} &
		\includegraphics[width=0.18\textwidth]{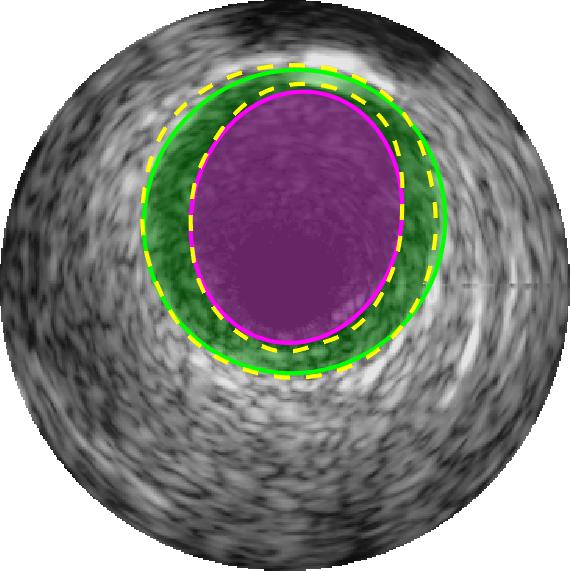} &
		\includegraphics[width=0.18\textwidth]{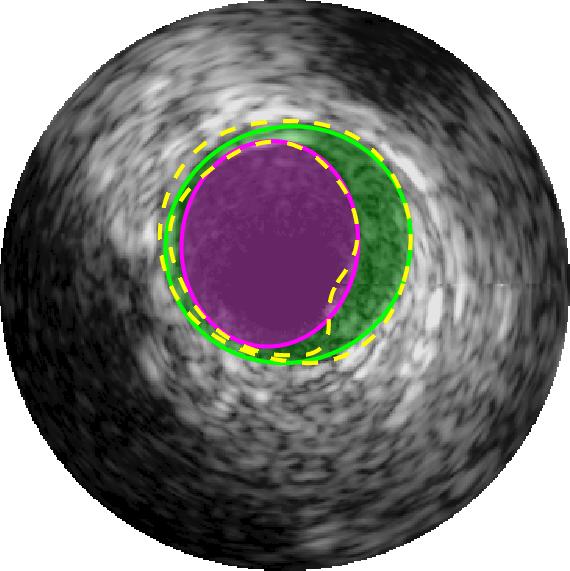} &	
		\includegraphics[width=0.18\textwidth]{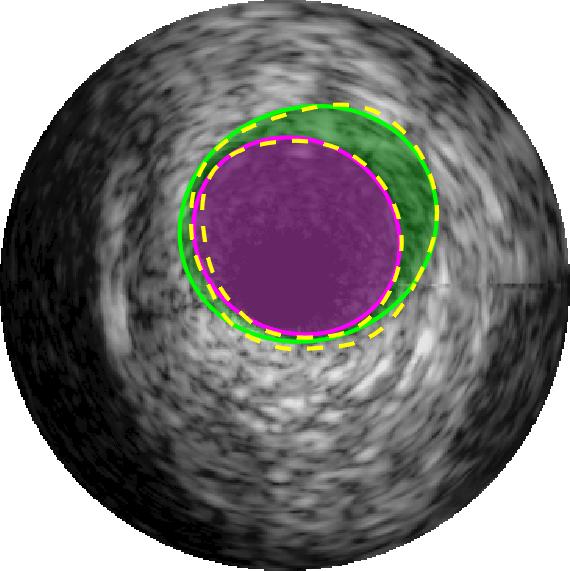} \\			  					  
	\end{tabular}
	\caption{Lumen and media segmentation results. Segmented lumen and media have been highlighted by magenta and green colours respectively. The yellow dashed lines illustrate the gold standard that have been delineated by four clinical experts \cite{balocco2014standardized}.}
	\label{fig::segmentationResults} 
\end{figure*}

\subsection{EREL Selection Results}
Qualitative evaluations are illustrated in Figure~\ref{fig::segmentationResults} and show the successful segmentation results of the proposed EREL selection strategy for 20 IVUS frames. The lumen areas are highlighted by the magenta colour while the media regions are green. Also, the manually annotated contours for both lumen and media are drawn as yellow dashed lines. As we can see, the chosen frames contain a variety of lumen and media morphologies. 

A detailed evaluation result and comparison with 9 recently published IVUS segmentation methods are reported in Table \ref{tbl::Generalcomparison} where the performance of the proposed EREL selection strategy in the presence of various artifacts is shown as well.

\begin{table*}\centering\footnotesize
		\caption{Performance of the proposed EREL selection strategy. Measures represent the mean and standard deviation evaluated on 435 frames of dataset B \cite{balocco2014standardized} and categorized based on the presence of a specific artifact in each frame. The evaluation measures are Jaccard Measure (JM), Hausdorff Distance (HD), and Percentage of Area Difference (PAD).}
	\begin{tabular}{|c|l@{}|c@{}|c@{}|c@{}|c@{}|c@{}|c@{}|}
		\cline{3-8} 
		\multicolumn{1}{c}{}&\multicolumn{1}{c|}{} & \multicolumn{3}{c}{\normalsize\textbf{Lumen}} \vline& \multicolumn{3}{c}{\normalsize\textbf{Media}} \vline\\
		\cline{3-8}
		\multicolumn{1}{c}{}& \multicolumn{1}{c|}{}&\textbf{HD} & \textbf{JM} & \textbf{PAD} &\textbf{HD} & \textbf{JM} & \textbf{Pad} \\ 
		\hline 
		\multirow{9}{*}{\parbox[c]{1.8cm}{\centering \scriptsize \textbf{General}\\ \textbf{Performance}}} 

		& Proposed Method                                      & 0.30 (0.20) & 0.87 (0.06) & 0.08 (0.09)  & 0.67 (0.54) & 0.77 (0.17) & 0.19 (0.18)\\ 
		&Unal et al. \cite{unal2008shape}                      & 0.47 (0.39) & 0.81 (0.12) & 0.14 (0.13)  & 0.64 (0.48) & 0.76 (0.13) & 0.21 (0.16)\\
		&Wang et al.\cite{balocco2014standardized}             & 0.51 (0.25) & 0.83 (0.08) & 0.14 (0.12)  & --          & --          &--\\
		&Destrempes et al.\cite{balocco2014standardized}       & 0.34 (0.14) & 0.88 (0.05) & 0.06 (0.05)  & 0.31 (0.12) & 0.91 (0.04) & 0.05 (0.04) \\
		&Downe et al. \cite{downe2008segmentation}             & 0.47 (0.22) & 0.77 (0.09) & 0.15 (0.12)  & 0.76 (0.48) & 0.74 (0.17) & 0.23 (0.19)\\
		&Alberti et al.\cite{balocco2014standardized}          & 0.46 (0.30) & 0.79 (0.08) & 0.16 (0.09)  & --          & --          &--\\
		&Ciompi et al. \cite{ciompi2012holimab}                & --          & --          &--            & 0.57 (0.39) & 0.84 (0.10) & 0.12 (0.12)\\
		&Mendizabal et al. \cite{mendizabal2013segmentation}   & 0.38 (0.26) & 0.84 (0.08) & 0.11 (0.12)  & --          & --          &--\\
		&Exarchos et al. \cite{balocco2014standardized}        & 0.42 (0.22) & 0.81 (0.09) & 0.11 (0.11)  & 0.60 (0.28) & 0.79 (0.11) & 0.19 (0.19)\\
		\hline
		\multirow{2}{*}{\centering \scriptsize \textbf{No Artifact}} 
		& Proposed Method  & 0.29 (0.17)& 0.88 (0.05) & 0.08 (0.07)   & 0.31 (0.23) & 0.89 (0.07) & 0.07 (0.08) \\
		& Lo Vercio et al. \cite{vercio2016assessment}            & --          & 0.83 (0.05) & 0.18 (0.06)  & --          & --          &--\\  

		\hline
		
		\multirow{8}{*}{\centering \scriptsize \textbf{Bifurcation}} 
		& Proposed Method                                      & 0.53 (0.34) & 0.79 (0.10) & 0.15 (0.17)   & 1.22 (0.45) & 0.57 (0.13) & 0.32 (0.19)\\ 
		&Unal et al. \cite{unal2008shape}                      & 0.65 (0.47) & 0.76 (0.14) & 0.18 (0.15)   & 0.57 (0.49) & 0.78 (0.13) & 0.19 (0.15)\\
		&Wang et al.\cite{balocco2014standardized}             & 0.54 (0.27) & 0.81 (0.11) & 0.14 (0.13)   & --          & --          &--\\
		&Destrempes et al.\cite{balocco2014standardized}       & 0.42 (0.18) & 0.85 (0.06) & 0.08 (0.06)   & 0.32 (0.13) & 0.91 (0.03) & 0.06 (0.04) \\
		&Downe et al. \cite{downe2008segmentation}             & 0.64 (0.27) & 0.70 (0.11) & 0.21 (0.15)   & 0.79 (0.53) & 0.71 (0.19) & 0.24 (0.21)\\
		&Alberti et al.\cite{balocco2014standardized}          & 0.61 (0.43) & 0.75 (0.10) & 0.20 (0.10)   & --          & --          &--\\
		&Ciompi et al. \cite{ciompi2012holimab}                & --          & --          &--             & 0.52 (0.29) & 0.85 (0.07) & 0.09 (0.07)\\
		&Mendizabal et al. \cite{mendizabal2013segmentation}   & 0.53 (0.36) & 0.79 (0.12) & 0.17 (0.18)   & --          & --          &--\\
		&Exarchos et al. \cite{balocco2014standardized}        & 0.47 (0.23) & 0.80 (0.09) & 0.10 (0.09)   & 0.63 (0.25) & 0.78 (0.11) & 0.23 (0.23)\\
		\hline

		\multirow{8}{*}{\centering \scriptsize \textbf{Side Vessels}} 
		& Proposed Method                                      & 0.24 (0.11) & 0.87 (0.05) & 0.06 (0.05)   & 0.74 (0.18) & 0.73 (0.60) & 0.21 (0.18)\\ 
		&Unal et al. \cite{unal2008shape}                      & 0.51 (0.39) & 0.79 (0.12) & 0.17 (0.14)   & 0.57 (0.39) & 0.78 (0.11) & 0.18 (0.12)\\
		&Wang et al.\cite{balocco2014standardized}             & 0.59 (0.23) & 0.80 (0.10) & 0.16 (0.13)   & --          & --          &--\\
		&Destrempes et al.\cite{balocco2014standardized}       & 0.36 (0.15) & 0.87 (0.04) & 0.07 (0.04)   & 0.31 (0.12) & 0.91 (0.04) & 0.04 (0.04) \\
		&Downe et al. \cite{downe2008segmentation}             & 0.46 (0.19) & 0.77 (0.08) & 0.15 (0.11)   & 0.76 (0.47) & 0.74 (0.16) & 0.22 (0.20)\\
		&Alberti et al.\cite{balocco2014standardized}          & 0.47 (0.24) & 0.79 (0.07) & 0.17 (0.09)   & --          & --          &--\\
		&Ciompi et al. \cite{ciompi2012holimab}                & --          & --          &--             & 0.53 (0.37) & 0.85 (0.09) & 0.10 (0.13)\\
		&Mendizabal et al. \cite{mendizabal2013segmentation}   & 0.38 (0.19) & 0.84 (0.07) & 0.11 (0.11)   & --          & --          &--\\
		&Exarchos et al. \cite{balocco2014standardized}        & 0.53 (0.24) & 0.77(0.09)  & 0.16 (0.12)   & 0.63 (0.31) & 0.78 (0.12) & 0.18 (0.16)\\
		\hline

		\multirow{8}{*}{\centering \scriptsize \textbf{Shadow}} 
		& Proposed Method                                      & 0.29 (0.20) & 0.86 (0.07) & 0.08 (0.09)  & 1.24 (0.39)  & 0.58 (0.13) & 0.37 (0.15)\\ 
		&Unal et al. \cite{unal2008shape}                      & 0.57 (0.39) & 0.78 (0.12) & 0.17 (0.12)  & 0.66 (0.50) & 0.77 (0.13) & 0.19 (0.15)\\
		&Wang et al.\cite{balocco2014standardized}             & 0.59 (0.27) & 0.81 (0.10) & 0.18 (0.16)  & --          & --          &--\\
		&Destrempes et al.\cite{balocco2014standardized}       & 0.39 (0.18) & 0.87 (0.05) & 0.06 (0.05)  & 0.33 (0.14) & 0.92 (0.03) & 0.05 (0.04) \\
		&Downe et al. \cite{downe2008segmentation}             & 0.55 (0.26) & 0.76 (0.11) & 0.14 (0.13)  & 0.77 (0.48) & 0.74 (0.16) & 0.22 (0.19)\\
		&Alberti et al.\cite{balocco2014standardized}          & 0.53 (0.29) & 0.78 (0.08) & 0.18 (0.09)  & --          & --          &--\\
		&Ciompi et al. \cite{ciompi2012holimab}                & --          & --          &--            & 0.58 (0.36) & 0.84 (0.09) & 0.11 (0.11)\\
		&Mendizabal et al. \cite{mendizabal2013segmentation}   & 0.43 (0.27) & 0.83 (0.09) & 0.12 (0.11)  & --          & --          &--\\
		&Exarchos et al. \cite{balocco2014standardized}        & 0.46 (0.19) & 0.80 (0.10) & 0.12 (0.12)  & 0.57 (0.28) & 0.82 (0.11) & 0.14 (0.17)\\
		\hline

	\end{tabular} 
	\label{tbl::Generalcomparison}
\end{table*}

\begin{figure*}[!t]
	\setlength\tabcolsep{2pt}
	\centering
	\begin{tabular}{ccccc}	
		\includegraphics[width=0.18\textwidth]{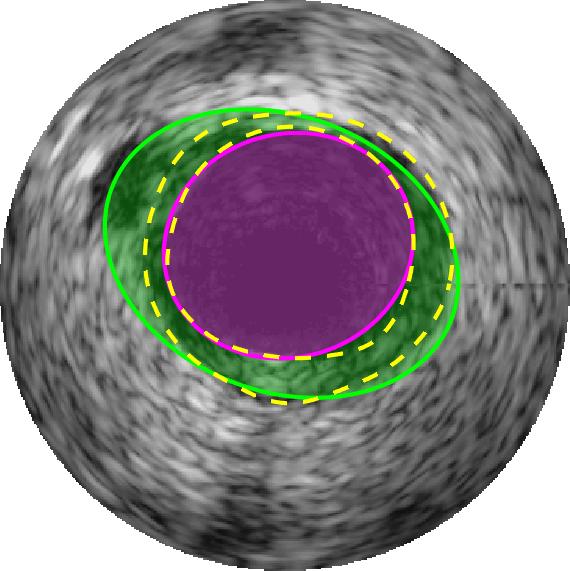} &
		\includegraphics[width=0.18\textwidth]{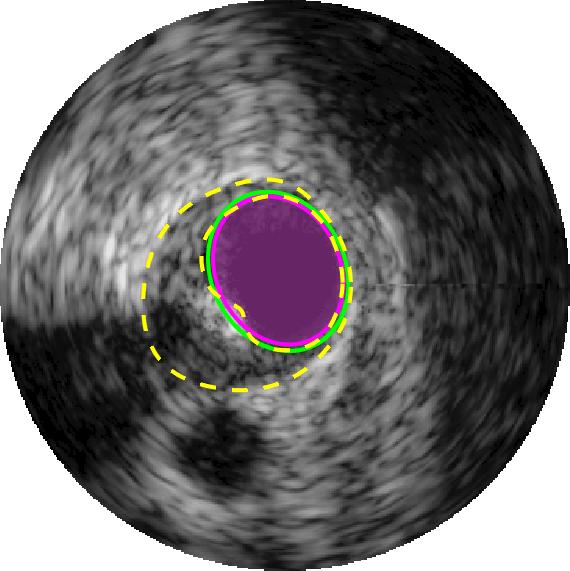} &
		\includegraphics[width=0.18\textwidth]{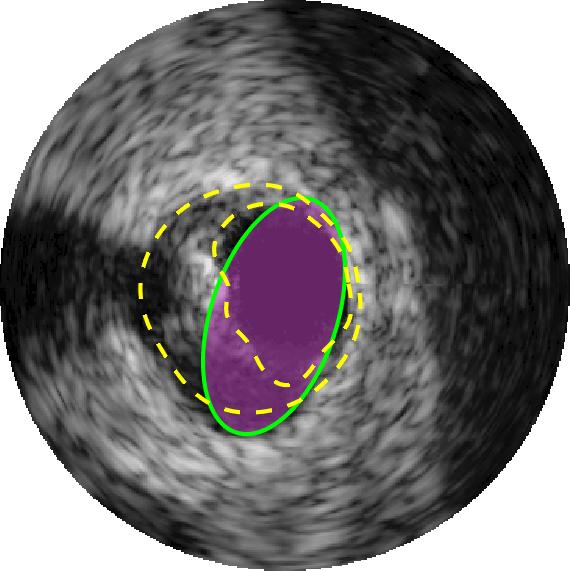} &
		\includegraphics[width=0.18\textwidth]{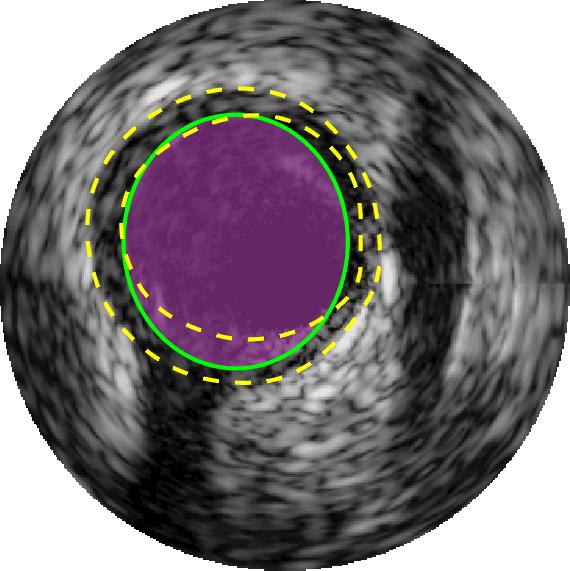} & 
		\includegraphics[width=0.18\textwidth]{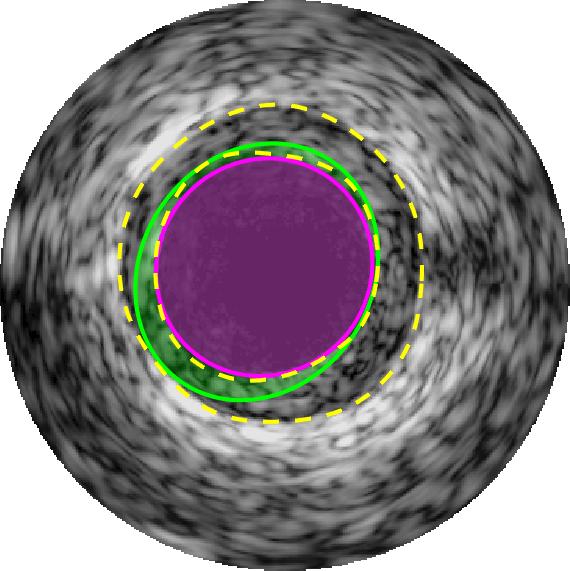} \\
		\includegraphics[width=0.18\textwidth]{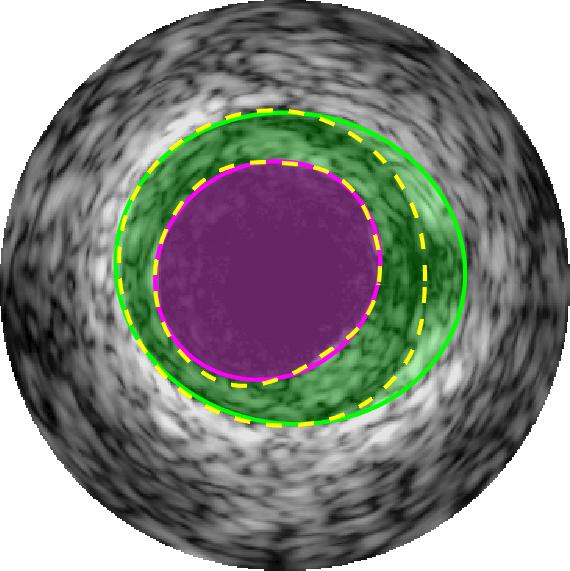} &
		\includegraphics[width=0.18\textwidth]{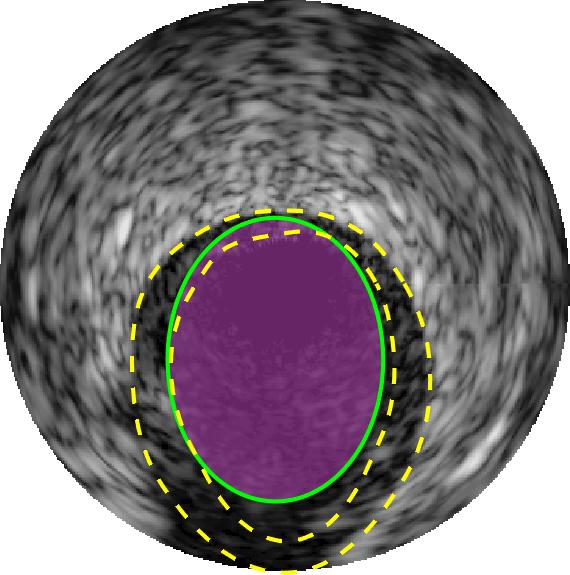} &	
		\includegraphics[width=0.18\textwidth]{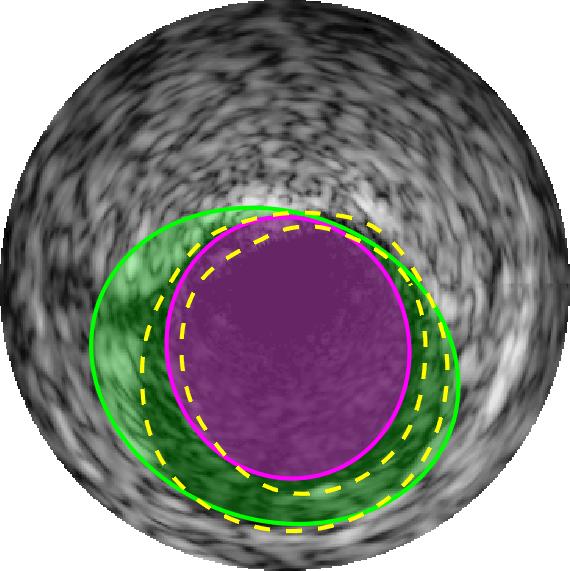} &			  
		\includegraphics[width=0.18\textwidth]{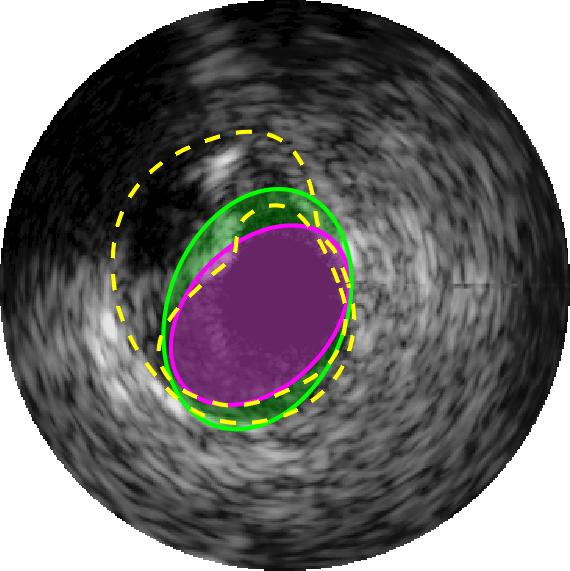} &
		\includegraphics[width=0.18\textwidth]{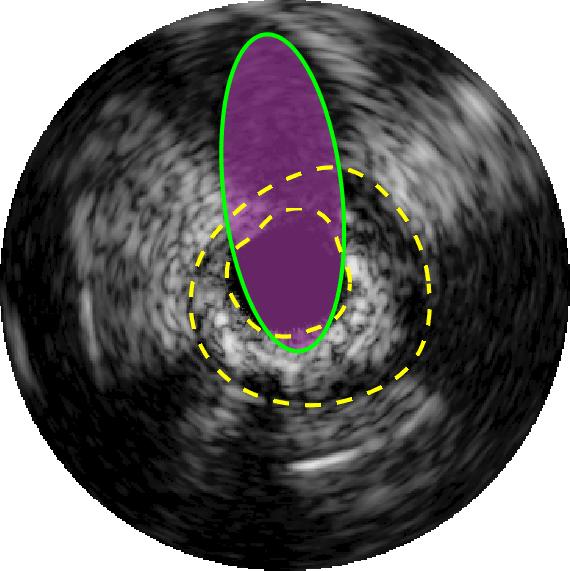} \\
	\end{tabular}
	\caption{Several inaccurate lumen and media segmentation results in the presence of various artifacts. Segmented lumen and media have been highlighted by magenta and green colours respectively. The yellow dashed lines illustrate the gold standard that have been delineated by four clinical experts \cite{balocco2014standardized}.}
	\label{fig::IncorrectSegmentationResults} 
\end{figure*}

\section{Discussion}\label{sec::discussion}
The present study is twofold. We first investigate whether the extracted ERELs \cite{faraji2015erel,faraji2015extremal} are proper regions to be considered as lumen and media delineations of 20 MHz Intravascular Ultrasound frames taken from the inside of the human coronary arteries. We quantitatively (Table~\ref{tbl::bestcase}) validate that very close candidates can be found from the small number (almost less than 50) of extracted ERELs for each IVUS frame. Specifically, the best case results demonstrate the average of the closest extracted EREL to the lumen and media among all 435 images of the dataset \cite{balocco2014standardized}. As the results of the best case study suggest, if we design a proper selection strategy we are able to find two regions out of the extracted ERELs that are very close to the actual lumen and media structure.

Secondly, we propose an ERELs selection strategy based on creating a vector that represents several textural characteristics of the regions and search for the regions which have higher textural stability scores. This method works very well when no artifact is present in the IVUS image or more accurately, when no major artifact is attached to the media. In this case, the average Hausdorff distance for both lumen and media is less than $0.3$ mm which shows that the proposed method works very well for images without any strong artifact. Furthermore, the general performance of our proposed method for lumen segmentation is superior to other available methods in terms of the average Hausdorff distance, which is $0.3$ mm. Likewise, our method is able to segment the lumen when the image contains shadows ($HD = 0.29$ mm) or side vessel ($HD = 0.24$ mm) artifacts, which is by far the lowest distance to the gold standard. 

If artifacts that are usually dark (low intensity value) regions, join the lumen and media regions, a type of leak will be created. This leak significantly increases the size and the mean intensity of the regions and generates irregular patterns that are very difficult to distinguish by the EREL extraction algorithm. Some of the inaccurate segmentations that are caused by the presence of the artifacts are shown in Figure~\ref{fig::IncorrectSegmentationResults}. As can be seen in Figure~\ref{fig::IncorrectSegmentationResults}, although the presence of the artifacts disrupts the detection of the media regions, most of the lumen regions can still be segmented accurately. This is also supported by the quantitative results presented in Table~\ref{tbl::Generalcomparison}. Evidently, even in the presence of artifacts such as bifurcations, side vessels, and shadows, the lumen segmentation performance remains high, though the accuracy of the media segmentation drops. In fact, even in the worst case condition of lumen segmentation that happens when the image contains bifurcation artifacts, the average Hausdorff distance to the gold standard is only $0.53$ mm which is still lower than most of the other methods. On the other hand, the media segmentation is more sensitive to the presence of artifacts. For instance, the Hausdorff metric for media in some cases increases to $1.22$ mm for bifurcation and to $1.24$ mm for shadow artifact. Accordingly, it can be concluded that the lumen segmentation is robust to the common artifacts of the IVUS images.   

The success of the proposed method depends strictly on the correct extraction of ERELs. In cases when continuous and clear boundaries do not exist between extremal regions, EREL cannot extract correct candidate regions for lumen and media. For example, in IVUS images taken with a 40~MHz catheter probe, some lumen pixels located on the boundaries adjacent to intima do not have clear connections to each other. A similar situation occurs for the boundaries between media and adventitia. This creates leakages that attach media region to the adventitia. So, the sequential process of EREL's enumeration connects pixels from two semantically different regions and constructs a big connected component that includes both regions. Eventually, the leakage obstructs EREL's region extraction capability for detecting $Q^+$ regions. This leakage can also be created in IVUS 20~MHz images by various artifacts that somehow attach the media regions to the adventitia as illustrated in Figure~\ref{fig::IncorrectSegmentationResults}. 

Apart from the above, our proposed method offers several benefits. Not only is our method automatic, but works also solely on the B-mode images (in contrast to some methods available in the literature that work on the polar space \cite{unal2008shape, taki2008automatic, mendizabal2013segmentation,zakeri2017automatic}) and, thus, it eliminates the need for transformation to the polar space. In addition, some methods \cite{sun2012parallel,cardinal2006intravascular} require the whole volume to be presented in order to segment the arterial walls. This makes it impossible to segment the lumen and media in online segmentation applications (during the in-vivo pullback procedures). On the other hand, our method delineates lumen and media contours using only the information available in the current frame and, hence, performs no redundant processing. Furthermore, due to the low computational complexity of the EREL \cite{faraji2015erel,faraji2015extremal}, every frame can be segmented in linear time based on the number of pixels in the frame. The average run time per frame in the proposed method over the test set of \cite{balocco2014standardized} is $0.19$ seconds.

\section{Conclusion}\label{sec::conclusion}
In this study, we investigated the suitability of a recently proposed region detector called \textit{Extremal Regions of Extremum Levels} for detecting the lumen and media regions. The results of our experiments reveal that among the ERELs extracted from a single IVUS frame there exist regions with very low difference from the manually annotated data. Therefore, we presented a region selection approach to automatically segment the arterial walls from Intravascular Ultrasound images taken with a 20 MHz catheter probe. We embedded the mean intensity value, the entropy, and the boundary length of the regions into a single feature vector representing the texture information of the region. Then, our selection method looks for regions with higher stability score. The region that is most stable at the beginning section of the texture vector is considered as the lumen. Similarly, the last region with a stable textural variation is labeled as media.

We evaluated our segmentation method on the test set of a standard publicly available IVUS dataset using three common evaluation metrics; namely, JM, HD, and PAD. Extensive quantitative comparisons show the high accuracy of our proposed method. In general, the average HD for our lumen segmentation results is $0.3$ mm which is the closest distance to the gold standard among all existing methods. In addition, when no artifact was present, our method segments both lumen and media achieving average HD lower than $0.29$ mm and $0.31$ mm, respectively. 

As reported in the best case results, among the extracted EREL regions, there exist regions very close to the actual structure of the arterial walls. In future work, we would like to devise a better selection strategy that chooses ERELs more accurately. 

Since IVUS is completely radiation free, unlike X-ray and CT, in the future we would also like to investigate the feasibility of combining the segmentation results from individual slices to obtain 3D views of collateral arteries.

\section{Acknowledgment}
We gratefully acknowledge the assistance of Prof. Simone Balocco at the University of Barcelona for providing the labeling information on the existing artifacts in the dataset.
\end{multicols}
\section*{References}

\bibliography{mybibfile}

\end{document}